%% file: arxiv.tex
\documentclass[10pt,twocolumn,letterpaper]{article}

\usepackage[pagenumbers]{iccv} 

\usepackage{bm}
\usepackage{xcolor}         

\usepackage{graphicx}
\usepackage{wrapfig}
\usepackage{cuted}
\usepackage{mathtools}

\usepackage{subcaption}

\usepackage{array}
\usepackage{booktabs}
\usepackage{multirow}

\definecolor{col1}{RGB}{255,255,255}
\definecolor{col2}{RGB}{255,255,255}
\definecolor{col3}{RGB}{255,255,255}

\newcommand{\model}{\textit{CoPart}\xspace}
\newcommand{\dataset}{\textit{PartVerse}\xspace}

\input{preamble}

\definecolor{iccvblue}{rgb}{0.21,0.49,0.74}
\usepackage[pagebackref,breaklinks,colorlinks,allcolors=iccvblue]{hyperref}
\definecolor{mypink}{RGB}{235,102,113}
\hypersetup{
    urlcolor = mypink,
}

\def\paperID{15118} 
\def\confName{ICCV}
\def\confYear{2025}

\title{From One to More: Contextual Part Latents for 3D Generation}

\author{
  Shaocong Dong$^{1}$\footnotemark[1], ~~Lihe Ding$^{2}$\footnotemark[1], 
  ~~Xiao Chen$^{2}$, ~~Yaokun Li$^{2}$, \\
  Yuxin Wang$^{1}$, ~~Yucheng Wang$^{1}$, ~~Qi Wang$^{1}$, ~~Jaehyeok Kim$^{1}$, \\
  Chenjian Gao$^{2}$, ~~Zhanpeng Huang$^{3}$, ~~Zibin Wang$^{3}$, \\
  ~~Tianfan Xue$^{2,4}$\footnotemark[2], ~~Dan Xu$^{1}$\footnotemark[2] \\
  {$^1$}HKUST 
  ~~~~~~{$^2$}CUHK 
  ~~~~~~{$^3$}SenseTime Research 
  ~~~~~~{$^4$}Shanghai AI 
  Laboratory \\
  {\tt\small\{sdongae, danxu\}@cse.ust.hk, \{dl023, tfxue\}@ie.cuhk.edu.hk}\\
  {\tt\small \{wangzb02\}@gmail.com, \{huangzhanpeng\}@sensetime.com}\\
}

\begin{document}
\maketitle
\begin{strip}
    \centering
    \vspace{-15mm}
    \includegraphics[width=\textwidth]{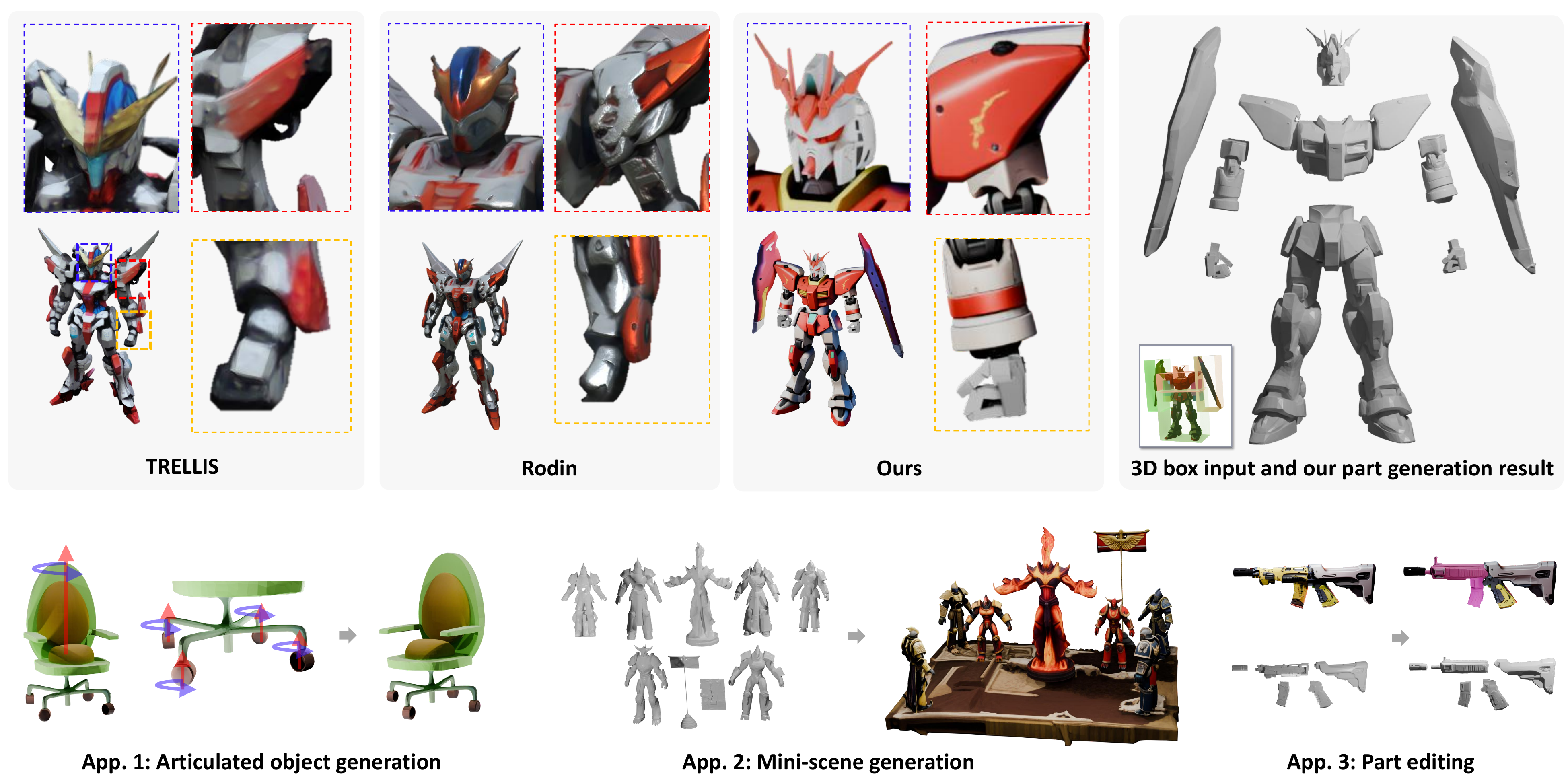}
    \vspace{-6mm} 
    \captionof{figure}{\model achieves high-quality part-based 3D generation and supports various applications.}
    \vspace{-1mm}
    \label{fig:teaser}
\end{strip}
\input{sec/0_abstract}
\input{sec/1_intro}
\input{sec/2_related}
\input{sec/3_framework}
\input{sec/4_applications}

\input{sec/5_dataset}
\input{sec/6_exp}

\input{sec/7_limitations}

{
    \small
    \bibliographystyle{ieeenat_fullname}
    \bibliography{main}
}


\end{document}

%% file: preamble.tex
\definecolor{darkred}{rgb}{0.7, 0.0, 0.0}


%% file: sec/0_abstract.tex
\begin{abstract}
To generate 3D objects, early research focused on multi-view-driven approaches relying solely on 2D renderings. Recently, the 3D native latent diffusion paradigm has demonstrated superior performance in 3D generation, because it fully leverages the geometric information provided in ground truth 3D data. Despite its fast development, 3D diffusion still faces three challenges. First, the majority of these methods represent a 3D object by one single latent, regardless of its complexity. This may lead to detail loss when generating 3D objects with multiple complicated parts. Second, most 3D assets are designed parts by parts, yet the current holistic latent representation overlooks the independence of these parts and their interrelationships, limiting the model's generative ability. Third, current methods rely on global conditions (e.g., text, image, point cloud) to control the generation process, lacking detailed controllability. Therefore, motivated by how 3D designers create a 3D object, we present a new part-based 3D generation framework, \model, which represents a 3D object with multiple contextual part latents and simultaneously generates coherent 3D parts. This part-based framework has several advantages, including: i) reduces the encoding burden of intricate objects by decomposing them into simpler parts, ii) facilitates part learning and part relationship modeling, and iii) naturally supports part-level control. Furthermore, to ensure the coherence of part latents and to harness the powerful priors from foundation models, we propose a novel mutual guidance strategy to fine-tune pre-trained diffusion models for joint part latent denoising. Benefiting from the part-based representation, we demonstrate that \model can support various applications including part-editing, articulated object generation, and mini-scene generation. Moreover, we collect a new large-scale 3D part dataset named Partverse from Objaverse through automatic mesh segmentation and subsequent human post-annotations. By training on the proposed dataset, \model achieves promising part-based 3D generation with high controllability. 
Project page: \url{https://copart3d.github.io}.
\end{abstract}

%% file: sec/1_intro.tex
\section{Introduction}
With the emergence of large-scale 3D datasets~\cite{deitke2022objaverse}, many
techniques have been proposed to convert the raw 3D data into different representations for generative modeling. Pioneering multi-view driven works~\cite{shi2023mvdream, liu2023syncdreamer} render 3D mesh into multi-view images and train multi-view diffusion models or large reconstruction models with only 2D image supervision. These methods can generate consistent multi-view images of 3D objects but are poor at recovering high-quality geometry since the accurate 3D shape supervision is omitted when converting mesh into multi-view images. Another 3D latent diffusion method CLAY~\cite{zhang2024clay} converts 3D meshes into latent tokens by a 3D VAE~\cite{kingma2013vae, zhang20233dshape2vecset} and trains a latent diffusion model. This method implicitly preserves previously overlooked geometric supervision through 3D occupancy auto-encoding, resulting in improved generation quality.

Despite all these advances of 3D latent diffusion, it still faces three challenges, making it still sub-optimal for 3D generation. First, the current methods treat intricate 3D meshes and simple ones equally, using the same number of tokens. However, the constrained representative ability of 3D VAE will inevitably cause information loss for complex data; and the unbalanced data distribution will make simpler geometries dominate the training process. Second, most 3D designers create complex 3D objects part by part, so they can spend more time adding detailed geometries for each part. On the contrary, state-of-the-art 3D generation algorithms neither utilize the part representations nor explicitly model the relationships between parts, limiting their ability to generate detailed and independent parts. For instance, when generating ``a person with a hat,'' most algorithms will fuse the head and hat together as a single object within a limited resolution of each local region, leading to low quality. However, users need two distinguishable detailed parts, especially for the face. 
Third, current methods utilize global conditions (e.g., text, image, point clouds) to control the generation process, which lacks detailed local controllability.



Based on these observations, we demonstrate that all these issues can be addressed through part-based 3D object representation and generation, as part-based modeling can i) naturally distribute complexity across individual parts, ii) efficiently learn part-level information from the data, and iii) provide detailed control at the part level. Therefore, we propose a novel part-based 3D generation framework, \textbf{\model}, which represents a 3D object with multiple \textbf{Co}ntextual \textbf{Part} latents and generates part latents by learning a joint distribution across diverse 3D data.


While 3D part generation has been explored previously, our solution offers a very distinct perspective. Majority of 3D part generation models either i) are restricted by PartNet~\cite{mo2019partnet} categories with limited generalizability~\cite{gao2019sdm, nakayama2023difffacto, koo2023salad}, or ii) adopt a ``top-down'' strategy~\cite{liu2024part123}, segmenting input images into part patches for individual reconstruction. The latter limits the model’s ability to leverage part information during training and depends heavily on segmentation quality. In contrast, \model adopts a ``bottom-up'' framework that directly generates coherent parts and leverage the diverse Objaverse~\cite{deitke2022objaverse} dataset, ensuring greater generalizability. 

Specifically, \model encodes each 3D part using a geometric token and an image token extracted from the part image, instead of one single global latent. This approach is beneficial for two reasons. First, each part has the complementary geometry and image tokens. Geometry tokens model the detailed shape, while the image tokens not only provide appearance information but also offer semantic cues for understanding part relationships. Second, decoupling geometry from image latents allows us to leverage the capabilities of pre-trained 3D and 2D autoencoders more effectively. For geometric tokens, since the part geometry is normally simpler than object geometry, they can be more efficiently encoded by a 3D part VAE~\cite{kingma2013vae}. For image tokens, since each part can be rendered in much higher resolution, 2D diffusion model can generate more detailed textures.  


To learn the distribution of both geometry and image tokens of each part, we finetune the diffusion models for 3D geometries~\cite{li2024craftsman} and 2D images~\cite{chen2023pixartalpha}, leveraging their pre-trained 
priors for better generation quality. To ensure both consistency between different parts and between geometric tokens and image tokens, we introduce a mutual guidance diffusion model, inspired by \cite{ding2024text}. The mutual guidance facilitates information exchange between different parts as well as between each part's geometric and image tokens, achieving both part consistency and geometry-image consistency. Furthermore, to eliminate the ambiguity of part order and effectively control the part generation using the input 3D bounding boxes, we propose a novel strategy to encode bounding box conditions to guide part generation. In this way, with the input of bounding boxes and text descriptions, we can generate high-quality 3D objects by decoding and assembling part latents, as shown in the first row of Fig.~\ref{fig:teaser}.

Collecting high-quality 3D part data for training is also non-trivial. One option is the PartNet dataset~\cite{mo2019partnet}, but it only contains 24 categories of objects and has poor textures, restricting model generalizability. Another option is Objaverse~\cite{deitke2022objaverse}, which offers more diversity but suffers from inconsistent part labels as different 3D designers prefer different ways to partition an object, often leading to over- or under-segmentation. To address this, we first employ a mesh segmentation pipeline to automatically decompose objects into reasonable parts. Then we manually conduct simple post-processing, including filtering low-quality data and grouping the over-segmented parts. Additionally, we utilize a multi-modal vision-language model~\cite{li2024mini} to generate text prompts for each part. In this way, we obtain a high-quality 3D part dataset with 91k parts for 12k objects.

With the proposed \model model trained on our 3D part dataset, it unlocks many various new applications, as shown in the bottom of Fig.~\ref{fig:teaser}. First, we can run structure diffusion~\cite{liu2024cage} to obtain bounding boxes and articulation information while using \model to generate parts, achieving novel articulated objects generation. Second, we can generate a mini-scene by considering each object in a scene as a part. Thirdly, we can achieve part-based 3D editing by resampling selected part latents. Experimental results also show that our method can generate high-quality 3D objects with more accurate parts compared with the previous works, and also support various applications as discussed above. 


%% file: sec/2_related.tex
\vspace{-2mm}
\section{Related Work}
\subsection{3D Generation}
In contrast to earlier category-specific 3D generation methods~\cite{Wu2016,Edward2017,Henzler2019,achlioptas2018learning,yang2019pointflow,Gao2019,Ibing2021,Chen2019,Park2019}, contemporary 3D generative models are capable of producing diverse 3D objects conditioned on text or images. DreamFusion~\cite{poole2022dreamfusion} and its subsequent works~\cite{wang2023prolificdreamer,lin2022magic3d,qian2023magic123} introduce the Score Distillation Sampling (SDS) loss to adapt 2D diffusion models for 3D generation. Multi-view diffusion approaches~\cite{shi2023mvdream,liu2023zero,liu2023syncdreamer} fine-tune 2D diffusion models to generate multi-view consistent images. Meanwhile, LRM~\cite{hong2023lrm} trains a large-scale reconstruction model to predict 3D radiance fields from a single image. Based on these 3D foundations, some work can achieve precise part and detail level modifications~\cite{dong2024interactive3d}. More recently, CLAY~\cite{zhang2024clay} and its follow-ups~\cite{li2024craftsman,yang2024tencent} directly train 3D-native diffusion models, achieving significantly improved performance.

\vspace{-2mm}
\subsection{Part Generation}
StructureNet~\cite{mo2019structurenet} uses a graph network to understand the relationship between parts while Grass~\cite{li2017grass} adopting recursive autoencoders for shape structures. DSG-Net~\cite{yang2022dsg} proposes disentangled structure and geometry for 3d generation. Other methods~\cite{genova2019learning, genova2020local} employ 3D Gaussian mixture to represent parts. SPAGHETTI~\cite{hertz2022spaghetti} and Neural Template~\cite{hui2022neural} train an auto-encoder to map 3D objects into a part-aware latent space, enabling part-aware editing. SALAD~\cite{koo2023salad} replaces the auto-encoder in SPAGHETTI with a diffusion model, achieving superior performance. DiffFacto~\cite{nakayama2023difffacto} learns a controllable part-based point cloud diffusion model. PartNeRF~\cite{tertikas2023generating} and NeuForm~\cite{lin2022neuform} also introduce part-based neural representations. While effective, these methods are limited by their reliance on category-specific part data, which restricts their generalizability. Part123~\cite{liu2024part123} leverages the powerful SAM~\cite{kirillov2023segany} model to segment multi-view images and perform part-aware reconstruction. Concurrent to our work, PartGen~\cite{chen2024partgen}, also adopts a "top-down" strategy by first segmenting parts from multi-view images and then performing part completion and reconstruction. However, this approach not only heavily depends on segmentation quality but also struggles with the limited information provided by small segmented patches, which constrains the quality of part reconstruction. In contrast, we propose a "bottom-up" strategy that directly learns the part distribution from diverse data and jointly generates coherent parts.

%% file: sec/3_framework.tex
\vspace{-2mm}
\section{Synchronized Part Latent Diffusion}

An effective part-based 3D generative model should be able to generate consistent parts with both high-quality geometry and appearance. However, this is non-trivial for the following three reasons. First, consistency between different parts is hard to ensure. Second, it is not easy to efficiently leverage limited part data to achieve high-quality part-based 3D generation. Third, simultaneously generating parts introduces ambiguity in part ordering.

\begin{figure*}[tbp]
\setlength{\abovecaptionskip}{-0.1cm}
\setlength{\belowcaptionskip}{-0.4cm}
\centering{\includegraphics[width=0.98\linewidth]{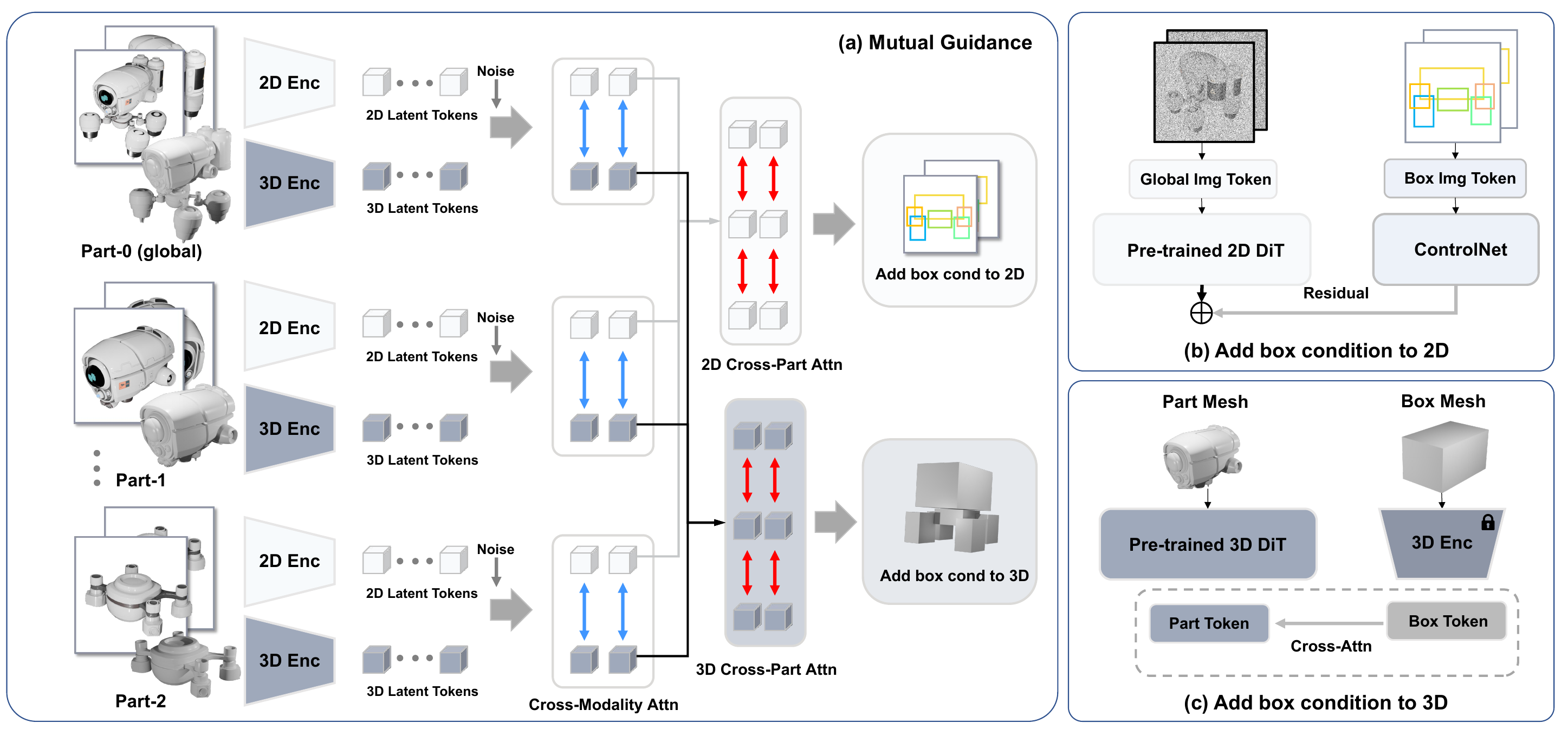}}
\caption{The framework of \model operates as follows: Gaussian noise is added to part image and geometric tokens extracted from the VAE, which are then fed into 3D and 2D denoisers. Mutual guidance (a) is introduced to facilitate information exchange between the 3D and 2D modalities (via Cross-Modality Attention) as well as between different parts (via Cross-Part Attention). Additionally, (b) the 3D bounding boxes are treated as cube meshes, and the extracted box tokens are injected into the 3D denoiser through cross-attention. Simultaneously, the boxes are rendered into 2D images and injected into the 2D denoiser via ControlNet. 
}
\label{fig:framework}
\end{figure*}

In this paper, we provide a synchronized part latent diffusion framework to address the above challenges as shown in Fig.~\ref{fig:framework}. In Sec.~\ref{encoding}, we first introduce our method to represent 3D objects using part latents. Next, in Sec.~\ref{guidance}, we propose an effective framework to synchronize part latent diffusion through mutual guidance, and fine-tune the part latent diffusion model from pre-trained foundation models for efficient part data utilization. Finally, in Sec.~\ref{box}, we discuss our approach to inject conditions to resolve the part order ambiguity and enhance controllability.

\vspace{-1mm}
\subsection{Part Representation Encoding}
\label{encoding}
To model the distribution of 3D parts, we decompose a 3D mesh $\mathcal{M}$ into part-based 3D representations that preserve both geometric and appearance information from the ground truth data. Our approach utilizes hybrid part latents to represent 3D texture parts through the combination of geometric tokens (encoded by a 3D part VAE) and image tokens (encoded by an image VAE), as detailed below. 

\vspace{-2mm}
\paragraph{Part Geometric Token Encoding.} 
Given a 3D part geometry $\mathcal{M}_{p}$ segmented from $\mathcal{M}$, we sample points $\bm{P} \in \mathbb{R}^{S \times 3}$ and their corresponding normals $\bm{Q} \in \mathbb{R}^{S \times 3}$ on the part mesh's surface, where $S=4096$ is the number of sampled points. Then a 3D part VAE encoder $\mathcal{E}_{3D}$ is used to extract 3D part geometric tokens $\bm{L}_{3D} = \mathcal{E}_{3D} (P, Q) \in \mathbb{R}^{T \times D}$, where $T$ and $D$ represent the token length and dimensions respectively. 
To enhance the part-level representation learning, we fine-tune the part VAE from a pre-trained holistic 3D VAE~\cite{li2024craftsman} using our part data. Additionally, we modify the VAE decoder $\mathcal{D}_{3D}$ from~\cite{li2024craftsman} to predict Flexicube~\cite{shen2023flexicubes} parameters, enabling differentiable rendering. More implementation details can be found in the supplementary material. These designs allow us to incorporate normal and depth rendering losses to supervise the VAE fine-tuning.

\vspace{-2mm}
\paragraph{Part Image Token Encoding.} 
To encode part appearance, we render the part mesh $\mathcal{M}_{p}$ into multi-view part-centric images $\{\bm{O}_k\}_{k=1}^{v}$. Using a pre-trained image VAE $\mathcal{E}_{2D}$~\cite{chen2023pixartalpha}, we obtain part image tokens: $ \bm{L}_{2D} = \{ \bm{F}_k | \bm{F}_k = \mathcal{E}_{2D} (O_k) \in \mathbb{R}^{T \times D} \}_{k=1}^{v} $ , where $v$ denotes the number of views, and $T$ and $D$ represent the token length and dimensions for each view respectively.

\vspace{-2mm}
\subsection{Synchronized Diffusion}
\label{diffusion}
\vspace{-1mm}
As introduced in Sec.~\ref{encoding}, a 3D object can be represented by $N$ part latents, comprising geometric tokens and image tokens: $\{ \bm{L}_{3D}^{p}, \bm{L}_{2D}^{p} \}_{p=1}^{N}$. 
To enable effective part generation, we leverage the powerful priors from pre-trained geometric and image diffusion models, and further fine-tune them with our part data.
Specifically, we fine-tune a pre-trained 3D latent diffusion~\cite{li2024craftsman} to generate geometric part latents $\bm{L}_{3D}^{p}$, while fine-tuning pre-trained image diffusion models~\cite{chen2023pixartalpha} to generate image part latents $\bm{L}_{2D}^{p}$.

To ensure consistent 3D object generation across all diffusion processes, we apply two types of synchronization. First is the inter-part synchronization, which ensures the part consistency. It synchronizes between parts $\bm{L}_{b}^{i}$ and $\bm{L}_{b}^{j}$, where $b \in \{ 3D, 2D \}$ denotes the modality and $i \ne j$ indicates different parts. Second is the intra-part synchronization between geometric and appearance representations $\bm{L}_{3D}^{i}$ and $\bm{L}_{2D}^{i}$ within the same $i$-th part, ensuring cross-modality (geometry and image modality) consistency. Details of both synchronization through guidance are described below.

\subsubsection{Mutual Guidance}
\label{guidance}
The inter-part synchronization design is inspired by BiDiff~\cite{ding2024text}, which synchronizes a 3D diffusion model and a 2D diffusion model through bidirectional guidance. We further extend this approach by adding mutual guidance between different part latents as well as between 3D and 2D modalities. The original bidirectional guidance, which replies on 2D-to-3D lifting and 3D-to-2D rendering, proves memory-intensive and inefficient for exchanging part-level information. Instead, we adopt a more effective implicit mutual guidance strategy that uses attention to exchange information between different parts and modalities (Fig.~\ref{fig:framework}) (a), as we use Transformer-based diffusion models~\cite{Peebles2022DiT}. Specifically, given noisy part latents $ \bm{L}_{3D}^{p, t}\  \text{and}\   \bm{L}_{2D}^{p, t} = \{ \bm{F}_k^{p, t} \}_{k=1}^{v}, p\!=\!1,...,N$ at diffusion timestep $t$, we define the intermediate features from the 3D and 2D diffusion as: 
\begin{equation}
\begin{split}
    \bm{\mathcal{G}}^{p} = \text{DiT} (\bm{L}_{3D}^{p, t},\  y, \ t), \ p=1,...,N
     \\ 
    \bm{\mathcal{I}}^{p} = \text{DiT} (\bm{L}_{2D}^{p, t},\  y, \ t), \ p=1,...,N,
\end{split}
\end{equation}
where $\bm{\mathcal{G}}^{p} \in \mathbb{R}^{T' \times D'}$ denotes the intermediate 3D features of $p$-th part, $\bm{\mathcal{I}}^{p} = \{ \bm{\mathcal{F}}_k^p \in \mathbb{R}^{T' \times D'} \}_{k=1}^{v} $ is the intermediate 2D features of $p$-th part conatining $v$ views, and $y$ represents additional conditions such as bounding boxes.

To ensure the inter-part consistency, we extend selected self-attention blocks in each modality to attend tokens from all parts:
\vspace{-2mm}
\begin{equation}
\begin{split}
    {\bm{\mathcal{G}}^{p}}' = \text{Attention}(\bm{\mathcal{G}}^{p}, \{ \bm{\mathcal{G}}^{i} \}_{i=1}^{N})
    \\
    {\bm{\mathcal{F}}_k^{p}}' = \text{Attention}(\bm{\mathcal{F}}_k^{p}, \{ \bm{\mathcal{F}}_k^{i} \}_{i=1}^{N}),
\end{split}
\label{eq:cross_modality}
\end{equation}
where Attention(query, key/value) is a standard attention block. This mechanism enables each part to be guided by other parts, facilitating inter-part mutual guidance and synchronization.

Similarly, to ensure cross-modality consistency, we add new attention blocks to exchange information between 3D and 2D features:
\begin{equation}
\begin{split}
    {\bm{\mathcal{G}}^{p}}' &= \bm{\mathcal{G}}^{p} + \text{LN}(\text{Attention}(\bm{\mathcal{G}}^{p}, \{ \bm{\mathcal{F}}_k^{p} \}_{k=1}^{v}))
    \\
    {\bm{\mathcal{F}}_k^{p}}' &= \bm{\mathcal{F}}_k^{p} +  \text{LN}(\text{Attention}(\bm{\mathcal{F}}_k^{p}, \bm{\mathcal{G}}^{p})),
\end{split}
\end{equation}
where LN() is a linear layer initialized by zeros for training stability.
Furthermore, to guarantee multi-view consistency for 2D branch, we also extend some self-attention blocks in 2D diffusion to attend tokens from other views of the same part:
\begin{equation}
    {\bm{\mathcal{F}}_k^{p}}' = \text{Attention}(\bm{\mathcal{F}}_k^{p}, \{ \bm{\mathcal{F}}_k^{p} \}_{k=1}^{v}).
    \label{eq:mv_attn}
\end{equation}

\begin{figure*}[tbp]
\setlength{\belowcaptionskip}{-0.4cm}
\centering{\includegraphics[width=\linewidth]{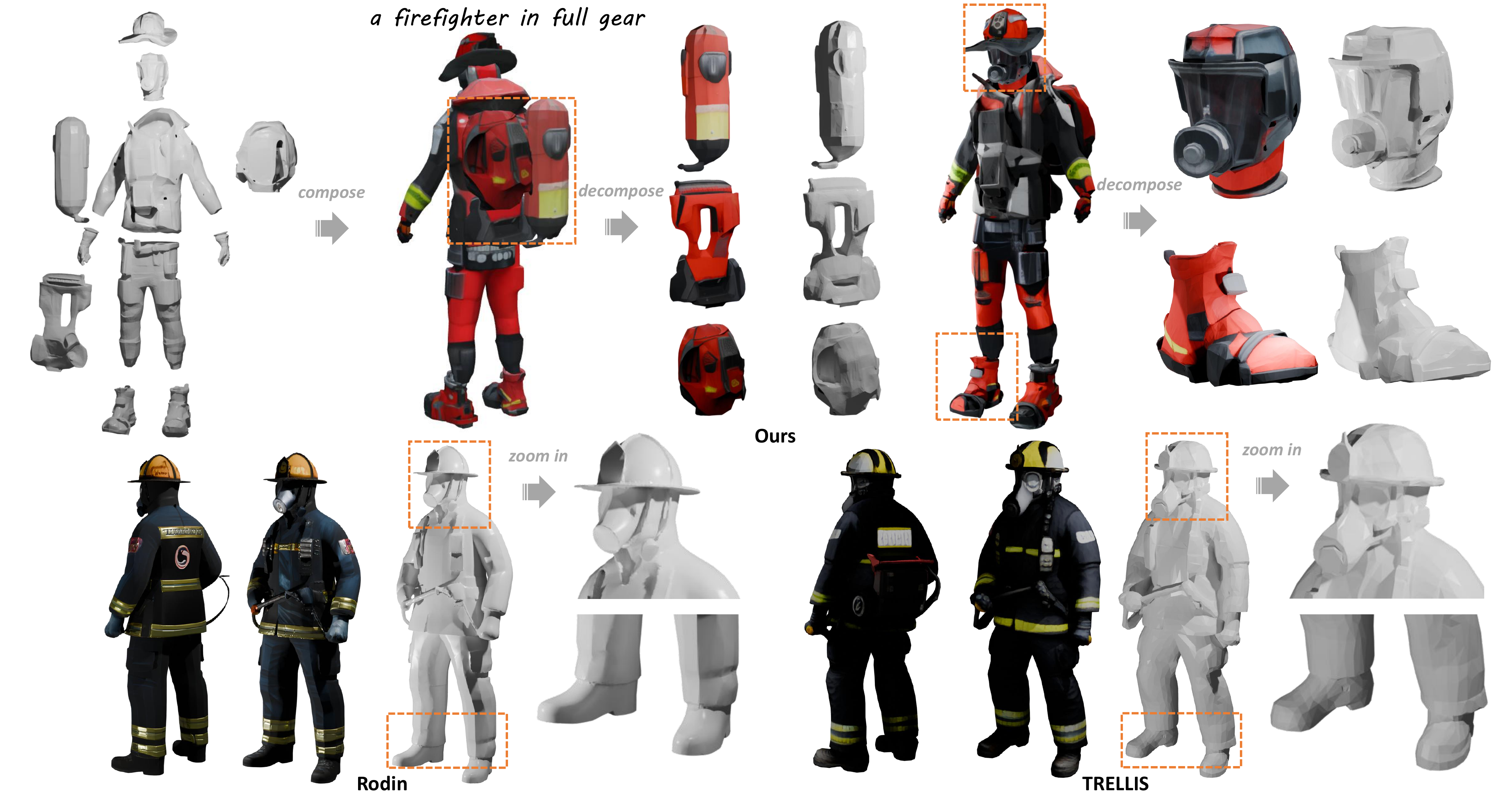}}
\vspace{-1em}
\caption{Comparison with state-of-the-art 3D generators. \model can generate detailed and independent 3D parts.}
\label{fig:comp_sota}
\vspace{-.3em}
\end{figure*}

\vspace{-2mm}
\subsubsection{Global Guidance} To further enhance inter-part consistency, we include a global branch for both 3D and 2D latent diffusions to jointly denoise holistic latents. This global branch functions as an additional ``global part'' that interacts with other part branches as mentioned in Sec.~\ref{guidance}. In particular, the global branch shares parameters with the part branch, distinguished only by concatenating learnable part-global embeddings to the text embeddings. 
This architecture enables the network to differentiate between part and global branches, and it further regularizes the fine-tuning process by maintaining global supervision, thereby preventing significant deviation from the original pre-trained weights.

\vspace{-1mm}
\subsubsection{Part Guidance Encoding}
\label{box}
One challenge of this design is the part order ambiguity. While we pre-define the part order during training, the network can alter this order during inference, creating a discrepancy between training and inference phases. To alleviate the part order ambiguity and enhance controllability, we introduce two conditions to both the 3D and 2D diffusion denoising processing.

First, we incorporate part-level text prompts to guide the network to distinguish different parts, thereby boosting local semantic controllability. Second, we introduce 3D bounding box conditions for each part as an additional constraint. A naive approach is to use MLPs (multi-layer perceptrons) to encode the coordinates of 3D bounding boxes and use the concatenation of coordinate embeddings with timestep embeddings as conditions. However, learning the correlation between embedded coordinates and actual 3D parts location is not easy.

To solve this challenge in 3D bounding box encoding, we propose a novel strategy to encode 3D bounding boxes by treating each box as a mesh with six surfaces, and then extracting box geometric latents through the pre-trained 3D mesh VAE encoder, as illustrated in Fig.~\ref{fig:framework} (c). The encoding can be written as
\vspace{-1mm}
\begin{equation}
    \bm{L}_{box}^{p} = \mathcal{E}_{3D}(\bm{P}_{box}^{p}, \bm{Q}_{box}^{p}), p=1,...N,
\end{equation}
where $\bm{P}_{box}^{p}$ and $\bm{Q}_{box}^{p}$ present the sampled points and normals on the $p$-th bounding box surfaces. In this way, we can encode the 3D bounding box information to the same latent space of the original geometric latents $\bm{\mathcal{G}}^{p}$, and simply added to the geometry latent through an additional cross-attention block:
\begin{equation}
    {\bm{\mathcal{G}}^{p}}' = \bm{\mathcal{G}}^{p} + \text{LN}(\text{Attention}(\bm{\mathcal{G}}^{p}, \bm{L}_{box}^{p})),
\end{equation}
where LN() denotes a linear layer initialized to zeros. 

For image tokens, we implement a dual-pathway approach to incorporate bounding boxes. First, we implicitly query bounding box information from 3D geometric tokens into image tokens through the cross-modal attention mechanism described in Eq.~\ref{eq:cross_modality}. Second, we encode 3D bounding boxes into the image latent space to guide the denoising of the global image branch.
Specifically, we render 3D boxes into 2D to generate multi-view bounding box wireframe images. These wireframe images are then encoded into latent tokens by an image VAE and integrated into the 2D denoiser through a lightweight ControlNet~\cite{zhang2023controlnet}, as illustrated in Fig.~\ref{fig:framework} (b). It is noteworthy that these bounding box 2D features are exclusively added to the global branch of the 2D diffusion model.

\vspace{-1mm}
\subsection{Refinement}
\label{sec:refine}
\vspace{-1mm}
After fine-tuning on the part dataset, the synchronized part latent diffusion is capable of understanding 3D parts and jointly generating consistent part geometric tokens and image tokens. These tokens can be decoded 
into part meshes and multi-view part-centric images by VAE decoders $\mathcal{D}_{3D}$ and $\mathcal{D}_{2D}$:
\begin{equation}
    {\mathcal{M}^{p}} = \mathcal{D}_{3D}({\bm{L}_{3D}^{p, 0}}), \ {\mathcal{O}^{p}} = \mathcal{D}_{2D}({\bm{L}_{2D}^{p, 0}}).
\end{equation}
To further improve quality, we leverage the 3D foundation model~\cite{xiang2024structured} as an additional enhancer, utilizing both the part images and geometry generated by our model. While the original pipeline of~\cite{xiang2024structured} takes a single image and a generated voxel as input, we found it incapable of understanding diverse 3D parts. Therefore, we modify this approach by replacing the voxels with our detailed part voxels extracted from the generated part meshes, thereby providing essential part geometry prior as follows:
\begin{equation}
    {\mathcal{M}^p}' = \mathcal{R}(\bm{O}^p, \bm{V}^p), \ \bm{V}^p = \text{Voxelize}(\mathcal{T}(\mathcal{M}^p)),
\end{equation}
where $\mathcal{R}$ represents the stage II enhancer, and $\mathcal{T}$ denotes a transformation to normalize $\mathcal{M}^p$ to $[-1, 1]$.
We can further enhance each part efficiently and assemble the parts by using the inverse transformation $\mathcal{T}^{-1}$: 
\vspace{-2mm}
\begin{equation}
    \mathcal{M}' = \{ \mathcal{T}^{-1}({\mathcal{M}^{p}}') \}_{p=1}^{N}.
\end{equation}
\vspace{-4mm}

\vspace{-4mm}
\subsection{Optimization Loss}
We supervise both the 3D and image branches of \model by the denoising loss in diffusion models:
\vspace{-2mm} 
\begin{equation}
{\mathclap{
\begin{split}
    Loss_{3D} \! &= \! \frac{1}{N} \! \sum_{p=1}^{N}( \mathbb{E}_{\bm{L}_{3D}^{p, t}, \bm{\epsilon}^{p}_{3d}, t }\| \bm{\epsilon}^{p}_{3d} \! - \! \mathcal{N}_{3d}(\bm{L}_{3D}^{p, t}, \bm{L}_{2D}^{p, t}, t)\|_{2}^{2} ),\\ 
    Loss_{2D} \! &= \! \frac{1}{N} \! \sum_{p=1}^{N}( \mathbb{E}_{\bm{L}_{2D}^{p, t}, \bm{\epsilon}^{p}_{2d}, t }\| \bm{\epsilon}^{p}_{2d} \! - \! \mathcal{N}_{2d}(\bm{L}_{3D}^{p, t},\bm{L}_{2D}^{p, t}, t)\|_{2}^{2} ),
\end{split}}
}
\end{equation}
where $\mathcal{N}$ represents the denoiser for 3D and 2D and $\bm{\epsilon}$ denotes Gaussian noise. 

%% file: sec/4_applications.tex

\begin{figure*}[htbp]
\setlength{\belowcaptionskip}{-0.35cm}
\centering{\includegraphics[width=\linewidth]{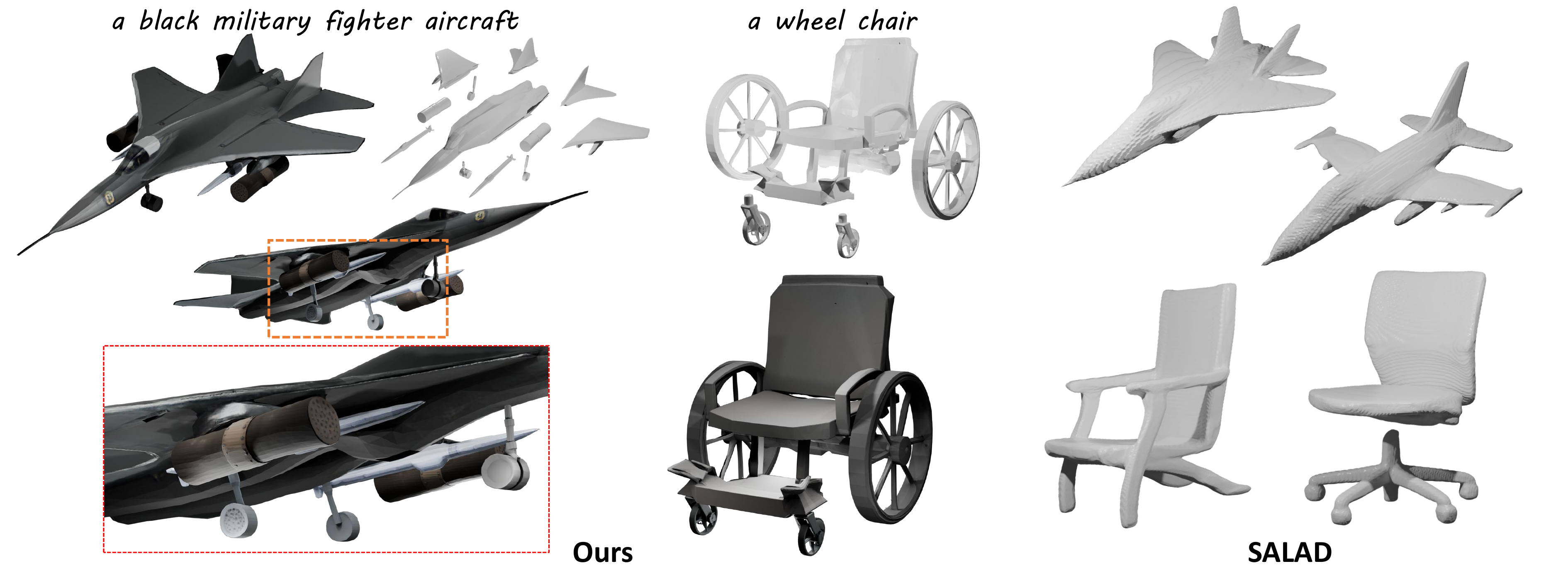}}
\vspace{-1em}
\caption{Comparison with part-based generator SALAD.}
\label{fig:comp_salad}
\vspace{-.3em}
\end{figure*}

\begin{figure*}[tbp]
\setlength{\belowcaptionskip}{-0.35cm}
\centering{\includegraphics[width=0.9\linewidth]{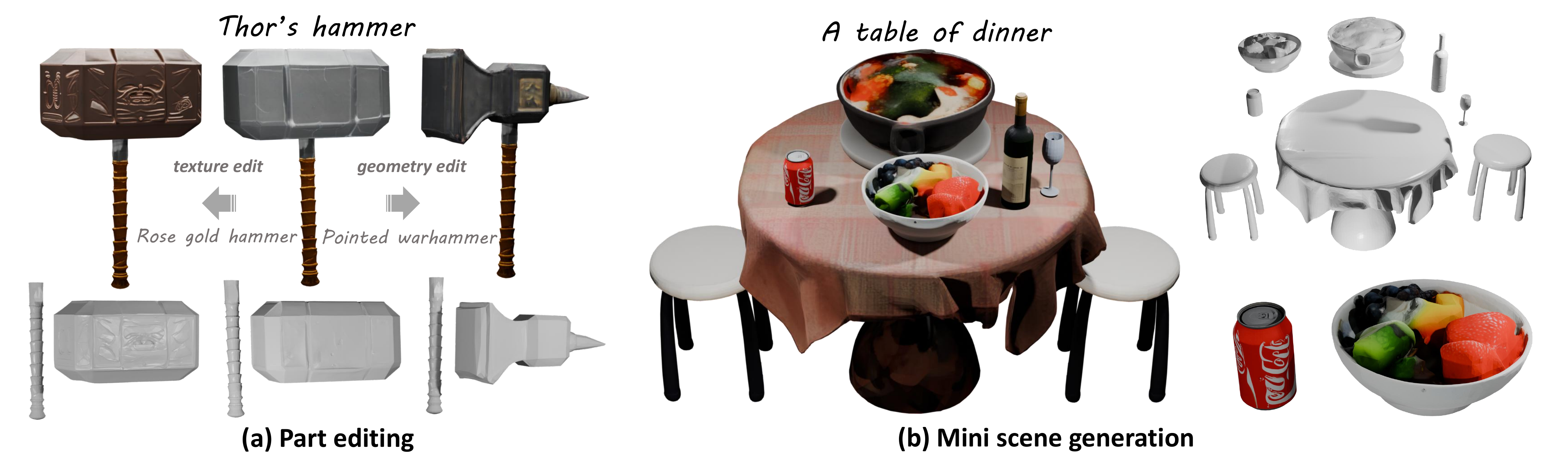}}
\vspace{-1em}
\caption{Qualitative results of part editing and mini scene generation.}
\label{fig:edit}
\vspace{-3mm}
\end{figure*}

\vspace{-2mm}
\section{Applications}
One major advantage of our part-based representation of \model is that it enables us to directly achieve various applications without further training. These applications include part-based editing (Sec.~\ref{sec:editing}), articulated object generation (Sec.~\ref{sec:articulate}), mini-scene generation and long part sequence sampling (Sec.~\ref{sec:mini_scene}).

\vspace{-1mm}
\subsection{Part-based Editing}
\label{sec:editing}
To enable selective part modification while keeping other parts unchanged, we design an inference-time resampling strategy. Specifically, given a mesh parts sequence $\{{\mathcal{M}^{p}}'\}_{p=1}^{N}$ either from \model sampling or segmented from an existing mesh, we denote the parts that need to be edited as $\{{\mathcal{M}^{p}}'\}_{p\in \mathcal{C}}$, where $\mathcal{C}$ is the selected index. To maintain the remaining parts unchanged during sampling while allowing them to provide information for new parts, we first encode the remaining part mesh back into contextual part latents:
\vspace{-2mm}
\begin{equation}
\begin{split}
    {\bm{L}_{3D}^{p, 0}}' = \  & \mathcal{E}_{3D}(\text{Sample}({\mathcal{M}^{p}}')) \\
    {\bm{L}_{2D}^{p, 0}}' = \ & \mathcal{E}_{2D}(\text{Render}({\mathcal{M}^{p}}')).
\end{split}
\end{equation}
Then during each timestep of the new editing sampling process, we directly replace the noisy part latents $\{ \bm{L}_{3D}^{p, t}, \ \bm{L}_{2D}^{p, t} \}_{p \notin \mathcal{C}}$ by adding noise to $\{ {\bm{L}_{3D}^{p, 0}}', \ {\bm{L}_{2D}^{p, 0}}' \}_{p \notin \mathcal{C}}$:
\vspace{-2mm}
\begin{equation}
\begin{split}
    \widehat{{\bm{L}_{3D}^{p, t}}} &= \sqrt{\overline{\alpha}_{t}} {\bm{L}_{3D}^{p, 0}}' + \sqrt{1 - \overline{\alpha}_{t}} \bm{\epsilon}_{3d}, p \notin \mathcal{C} \\
    \widehat{\bm{L}_{2D}^{p, t}} &= \sqrt{\overline{\alpha}_{t}} {\bm{L}_{2D}^{p, 0}}' + \sqrt{1 - \overline{\alpha}_{t}} \bm{\epsilon}_{2d}, p \notin \mathcal{C},
\end{split}
\end{equation}
where $\overline{\alpha}_{t}$ is noise schedule and $\bm{\epsilon}$ is random Gaussian noise for 3D or 2D. Thus, we can sample additional part latents from pure Gaussian noise while incorporating information from the fixed one by:
\begin{equation}
    {{\bm{L}_{3D}^{p,t-1}}\!,\!{\bm{L}_{2D}^{p, t-1}}} \!=\! \mathcal{N}(\{\widehat{\bm{L}_{3D}^{p, t}}, \widehat{\bm{L}_{2D}^{p, t}}\}_{p \notin \mathcal{C}},\!\{\bm{L}_{3D}^{p, t},\!\bm{L}_{2D}^{p, t}\}_{p \in \mathcal{C}}),
\end{equation}
where $\mathcal{N}$ is the diffusion denoiser. 
In this way, we can modify the text prompts to edit the selected parts as shown in Fig.~\ref{fig:edit} (a).

\vspace{-2mm}
\subsection{Articulated Object Generation}
\label{sec:articulate}
The generation of articulated objects involves two key components: i) articulation generation, which includes the bounding boxes indicating the position of each part, and ii) part generation. To achieve this, we leverage an off-the-shelf method~\cite{liu2024cage} to generate the bounding boxes along with their articulation relationships. Subsequently, we utilize \model to populate each bounding box with coherent parts based on text prompts. This approach enables the creation of novel articulated objects, such as an avocado swivel chair (Fig.~\ref{fig:teaser}), which cannot be realized using previous holistic generation methods. Additionally, the part-based generation approach provides the flexibility to manually define articulation information for each part. \emph{Please refer to the supplementary video for the 
visualization of generated articulated objects}. 

\vspace{-1mm}
\subsection{Mini-scene and Long Sequence Generation}
\label{sec:mini_scene}
\vspace{-2mm}
Our approach extends naturally to mini-scene generation, as scenes can be represented as layouts of bounding boxes, aligning with our box-guided part generation where each part corresponds to an object. Our training data includes mini-scenes, enabling direct scene generation through specified boxes and text (Fig.~\ref{fig:edit} (b)). For complex objects requiring long part sequences, GPU memory constraints during training limit the maximum number of parts $N$=8. We address it by adopting the strategy from Sec.~\ref{sec:editing}: fixing some sampled part latents while replacing others with new box conditions sampled from Gaussian noise. This enables the generation of longer sequences without memory issues. More details can be found in the supplementary materials.

%% file: sec/5_dataset.tex
\begin{figure*}[htbp]
\setlength{\abovecaptionskip}{-0.15cm}
\setlength{\belowcaptionskip}{-0.15cm}
\centering{\includegraphics[width=\linewidth]{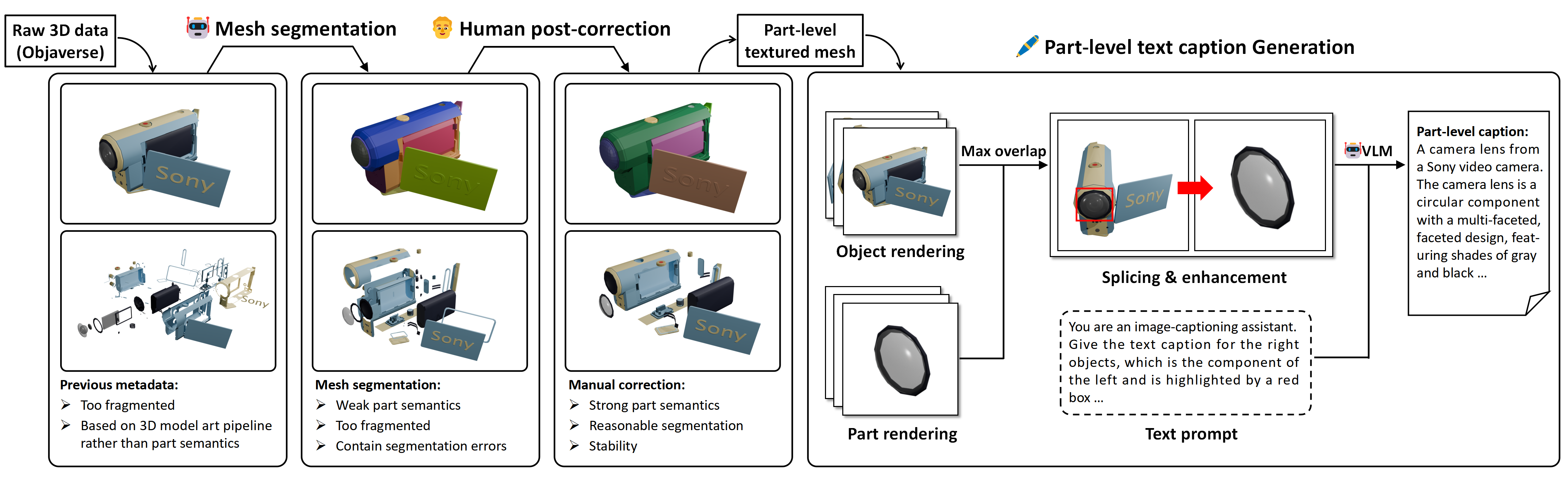}}
\vspace{-1em}
\caption{PartVerse dataset processing pipeline. We follow the pipeline of ``raw data - mesh segment algorithm - human post correction - generate text caption" to produce part-level data.}
\label{fig:dataset_pipeline}
\end{figure*}

\begin{table*}[htbp]
\centering
\setlength{\abovecaptionskip}{0.0cm}
\caption{Quantitative comparison with SOTA methods. CLIP (N-T) and CLIP (I-T)~\cite{radford2021clip,zhang2024clay} gauge the geometry alignment of normal maps with the input text and similarity of render images with the input text, respectively. In addition, ULIP-T~\cite{xue2023ulip} was also experimented and user-prefer study were conducted. $\dag$ Our method takes 12s when the number of parts is one.}
\label{tab:quantitive}
\setlength{\arrayrulewidth}{1pt} 
\begin{tabular}
{@{}l| ccc | ccc | c @{}}
\toprule[1.5pt]
\multirow{2}*{\textbf{Method}} & \multicolumn{3}{c|}{\textbf{Whole-aware}} & \multicolumn{3}{c|}{\textbf{Part-aware}} & \multirow{2}*{\textbf{Time}} \\ 
& CLIP (N-T) & CLIP (I-T) & ULIP-T & CLIP (N-T) & CLIP (I-T) & ULIP-T & \\
\midrule[0.8pt]
\midrule[0.8pt]
Shap-E~\cite{jun2023shap}  & 0.1546 & 0.1607 & 0.1054 & 0.0875 & 0.1043 & 0.0795  & 3s   \\
Unique3D~\cite{wu2024unique3d} & 0.1865 & 0.2037 & 0.1528 & 0.1062 & 0.1380 & 0.1032     & 16s    \\
CraftMan~\cite{li2024craftsman} & 0.1887 & 0.1966 & 0.1476 & 0.1026 & 0.1271 & 0.0997     & 8s   \\
Rodin~\cite{hyper3d.ai} & \underline{0.2042} & \textbf{0.2416} & \textbf{0.1785} & \underline{0.1425} & \underline{0.1571} & \underline{0.1244}  & -  \\
Trellis~\cite{xiang2024structured} & \textbf{0.2071} & 0.2360 & \underline{0.1751} & 0.1274 & 0.1455 & 0.1119  & 10s  \\
Ours          & 0.2010 & \underline{0.2387} & 0.1743 & \textbf{0.1607} & \textbf{0.1768} & \textbf{0.1355}    & 65s$^{\dag}$ \\
\bottomrule[1.5pt]
\end{tabular}
\vspace{-5mm}
\end{table*}

\vspace{-3pt}
\section{PartVerse Dataset}
\vspace{-3pt}
To enhance the generalizability of \model, we introduce \dataset, a new diverse 3D part dataset comprising 91k high-quality parts from 12k objects with detailed text descriptions. Unlike previous part datasets such as PartNet~\cite{mo2019partnet} which only contains \textbf{24} categories of daily objects, our \dataset, curated and annotated from Objaverse~\cite{deitke2022objaverse}, exhibits enhanced diversity in \textbf{175} categories and realistic textures, significantly improving the model's ability to generate high-quality 3D parts. As shown in Fig.~\ref{fig:dataset_pipeline}, we provide an overview of the data collection and annotation pipeline used to create Partverse. 

\noindent\textbf{Automatic Mesh Segmentation.} 3D artists follow a specific modeling pipeline when creating 3D objects, typically creating them part by part. We can restore this information from the raw 3D data. However, the original modeling steps might not always match semantic part boundaries - for instance, some artists might create textured surfaces as separate elements. A pre-labeling algorithm based on SAM-2~\cite{ravi2024sam2} and Samesh~\cite{tang2024segment} was constructed using 3D model creation priors combined with semantic segmentation, which balances the mesh faces connectivity and visual semantic during segmentation. This algorithm can allow us to preliminarily obtain semantic parts. We specifically adjusted parameters to favor over-segmentation rather than under-segmentation, since it's easier for annotators to merge extra segments than to split insufficient ones.

\noindent\textbf{Human post-annotation.} After initial segmentation, human reviewers first remove low-quality data including overly complex or unsuitable 3D objects for splitting. Using a Blender-based~\cite{Blender} annotation platform, they then refine the segmentation by merging over-segmented or splitting under-segmented parts according to guidelines: ensuring clear part semantics and maximizing symmetry in part distribution. Based on these annotations, complete textured objects are finally split into individual part objects with textures preserved.

\noindent\textbf{Part-level text caption.} We also created textual descriptions for each part, covering appearance features, shape characteristics, and part-whole relationships. The process starts by rendering multi-view images of both complete objects and individual parts. We select the view showing maximum visible overlap between part and whole, then combine the object image with the highlighted part using a bounding box. This composite image is input to a vision-language model (VLM) to generate descriptions of the component and its relationship to the complete object.

%% file: sec/6_exp.tex
\vspace{-2mm}
\section{Experimental Evaluation}
\vspace{-2mm}

This section presents our experimental results. For better visualization, please refer to our supplementary video.

\noindent\textbf{Implementation details.}
We initialize our 2D and 3D denoisers using Pixart-$\alpha$~\cite{chen2023pixartalpha} and CraftMan~\cite{li2024craftsman}, and fine-tune them on PartVerse dataset with 4 NVIDIA A100 GPUs. We set the batch size to 1 and limit the maximum part number to 8. We also perform random selecting or padding during training. Moreover, we adopt a progressive training strategy, which is detailed in the supplementary material.

\noindent\textbf{Comparison with state-of-the-art 3D generators.}
We compare \model with leading holistic 3D generators in Fig.~\ref{fig:comp_sota}. More results can be found in the supplementary materials.
It is evident that \model outperforms state-of-the-art methods, particularly in the quality of small parts, owing to its part-based representation. 

\setlength{\columnsep}{12pt}
\begin{wraptable}{r}{4.4cm}
  \setlength\tabcolsep{1pt}
  \renewcommand\arraystretch{1.0}
  \setlength{\abovecaptionskip}{-0.0mm}  
  \setlength{\belowcaptionskip}{8pt}
  \centering
  \vspace{-7mm}
  \caption{User study (\%preference).}
  \label{tab:user}
  \footnotesize
    \begin{tabular}{c c c}
    \toprule
    Method & Whole-aware & Part-aware \\
    \midrule
    Rodin~\cite{hyper3d.ai} & 33.3\% & 25.5\% \\
    PartGen~\cite{chen2024partgen} & 11.8\% & 13.7\% \\
    Ours &  \textbf{54.9}\% & \textbf{60.8\%} \\
    \bottomrule
    \end{tabular}
    \vspace{-4mm}
\end{wraptable}
\noindent\textbf{Quantitative comparison.} As shown in Tab.~\ref{tab:quantitive}, we conduct quantitative comparisons following~\cite{zhang2024clay}. For the Image-to-3D model, we first generate corresponding images from given texts by Flux.1~\cite{yang2024flux}. Different from~\cite{zhang2024clay}, half of the test cases we use are part-aware, such as ``a rifle stock". This is reasonable, since a truly general 3D generation model should be able to handle all types. In addition, ULIP~\cite{xue2023ulip} was used in evaluation. 
As shown in \cref{tab:user}, we also conducted a user study with 51 diverse participants from different professions by collecting their preferences for generating textured mesh. These results highlights the advantages of our part-based generation approach in producing decomposable and high-quality 3D assets. More details can be found in the supplementary materials.


\noindent\textbf{Comparison with part-based generation methods.}
We compare \model with the accessible state-of-the-art part generator SALAD~\cite{koo2023salad}. Fig.~\ref{fig:comp_salad} shows that \model can generate diverse objects with detailed parts while SALAD is constrained to generate shapes in PartNet distribution. More comparisons can be found in the supplementary materials.

\begin{figure}[htbp]
  \vspace{-4mm}
  \setlength{\abovecaptionskip}{0cm}
  \setlength{\belowcaptionskip}{-0.3cm}
  \centering
  \includegraphics[width=\linewidth]{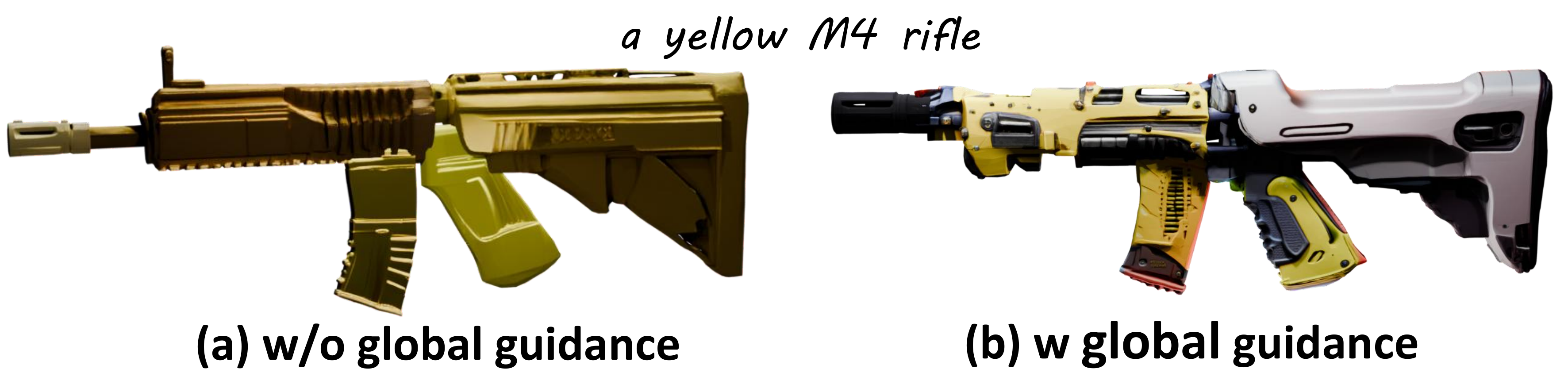}
  \caption{Ablation of global guidance.}
  \label{fig:global_ab}
\end{figure}

\noindent\textbf{Ablation study of global guidance.}
We ablate the effect of global guidance in Fig.~\ref{fig:global_ab}. The results demonstrate that global guidance significantly enhances part coherence, especially in appearance.

\begin{figure}[htbp]
  \setlength{\abovecaptionskip}{0.1cm}
  \setlength{\belowcaptionskip}{-0.4cm}
  \centering
  \includegraphics[width=\linewidth]{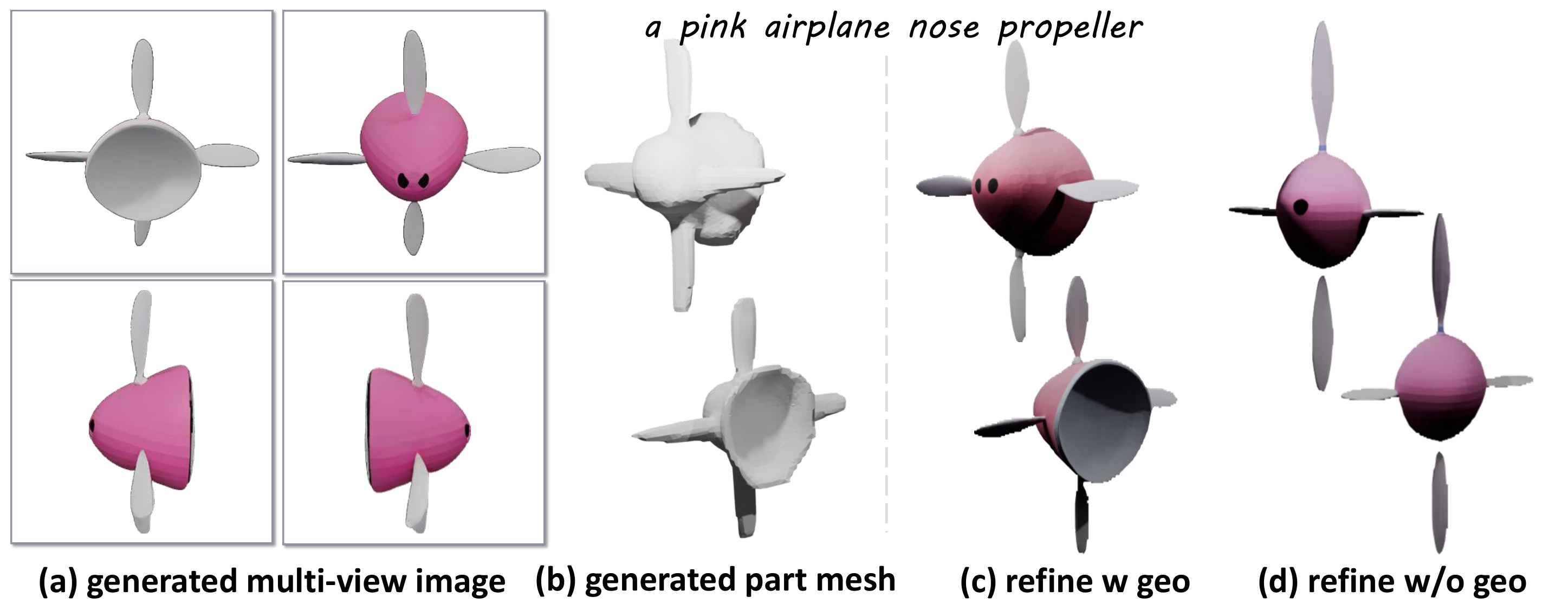}
  \caption{Ablation of refinement module.}
  \label{fig:refine_ab}
\end{figure}

\noindent\textbf{Ablation study of refinement.}
We ablate the effect of refinement (Sec.~\ref{sec:refine}), as shown in Fig.~\ref{fig:refine_ab}. The results show that providing both part images (Fig.~\ref{fig:refine_ab} (a)) and geometries (Fig.~\ref{fig:refine_ab} (b)) is essential for the enhancer to accurately comprehend part shapes. By integrating both modalities, the enhancer effectively optimizes the parts and enhances overall performance.

\vspace{-2mm}
\section{Discussion of Part Assembly}
\vspace{-2mm}
As depicted in Sec.~\ref{sec:refine}, the part latents are decoded in absolute positions and then normalized to perform refinement. After an inverse-normalization, the refined part meshes can be relocated to the 3D bonding boxes for assembling. Our precise strategy for injecting bounding box information generally ensures effective combinations of part meshes. However, assembly errors, such as ``mesh clipping", can occur when the user provides incorrect bounding boxes. We will explore techniques to optimize user-provided bounding boxes in future work.

%% file: sec/7_limitations.tex

\vspace{-2mm}
\section{Conclusion}
\vspace{-1mm}
We present \model for high-quality and diverse 3D part generation. Specifically, we utilize mutual guidance to ensure coherent part latent denoising and introduce 3D box conditions to eliminate part ambiguity. Furthermore, a larger scale 3D part-aware dataset is firstly collected from Objaverse, which can be widely used for various tasks. 
Our method outperforms SoTA results. 
We also discuss the limitations of our method in the supplementary material.